\documentclass[conference]{IEEEtran}
\IEEEoverridecommandlockouts
\usepackage{cite}
\usepackage{amsmath,amssymb,amsfonts}
\usepackage{graphicx}
\usepackage{textcomp}
\usepackage{hyperref}
\usepackage{booktabs}
\usepackage{multirow}
\usepackage{array,multirow}
\usepackage[dvipsnames]{xcolor}
\usepackage[T1]{fontenc}
\usepackage{tikz}
\usepackage{amsmath, amsfonts}

\def\BibTeX{{\rm B\kern-.05em{\sc i\kern-.025em b}\kern-.08em
    T\kern-.1667em\lower.7ex\hbox{E}\kern-.125emX}}

\newcolumntype{R}[1]{>{\raggedleft\arraybackslash }b{#1}}
\newcolumntype{L}[1]{>{\raggedright\arraybackslash }b{#1}}
\newcolumntype{C}[1]{>{\centering\arraybackslash }b{#1}}
\usetikzlibrary{arrows.meta} 
\def\nbVideo{\ $60$ } 
\def\nbSegments{\ $318$ } 
\def\nbWords{\ $2485$ }   %
\def\nbArabBertFeautures{\ $768$ } 
\def\nbAcousticFeatures{\ $52$ }   
\def\nbVisualFeatures{\ $45$ } 
\begin{document}

\setlength{\arrayrulewidth}{0.4mm}
\renewcommand{\arraystretch}{1.5}
 \makeatletter
    \newcommand{\linebreakand}{%
      \end{@IEEEauthorhalign}
      \hfill\mbox{}\par
      \mbox{}\hfill\begin{@IEEEauthorhalign}
    }
    \makeatother
\newcommand{\mail}[1]{\href{mailto:#1}{\texttt{#1}}} 

\title{Towards Arabic Multimodal Dataset  for   Sentiment Analysis

\thanks{This research is performed under the MESRS Project  PRFU: C00L07N030120220002 }
}

\author{
\IEEEauthorblockN{Abdelhamid Haouhat}
\IEEEauthorblockA{\textit{Lab. d'Informatique et de Mathématiques} \\
\textit{Université Amar Telidji, Laghouat, Alg\'erie }\\
\mail{a.haouhat@lagh-univ.dz}}
\and

\IEEEauthorblockN{Slimane Bellaouar }
\IEEEauthorblockA{\textit{Dept. of Mathematics and Computer Science} \\
\textit{Lab. des Mathématiques et Sciences Appliquées (LMSA)}\\
\textit{Université de Ghardaia, Alg\'erie} \\
\mail{bellaouar.slimane@univ-ghardaia.dz}}

\linebreakand
\IEEEauthorblockN{Attia Nehar}
\IEEEauthorblockA{
\textit{Universit\'e Ziane Achour} - Djelfa, Alg\'erie\\
\textit{Lab. d'Informatique et de Mathématiques (LIM)}\\
\textit{Université Amar Telidji, Laghouat, Alg\'erie} \\
\mail{neharattia@univ-djelfa.dz}}
\and
\IEEEauthorblockN{Hadda Cherroun}
\IEEEauthorblockA{\textit{Lab. d'Informatique et de Mathématiques} \\
\textit{Université Amar Telidji, Laghouat, Alg\'erie }\\
\mail{hadda\_cherroun@lagh-univ.dz}}
}

\maketitle
\begin{abstract}
Multimodal Sentiment Analysis (MSA) has recently become a
centric research direction for many real-world applications. This proliferation is due to the fact that opinions are central to almost all human activities and are key influencers of our behaviors. In addition, the recent deployment of Deep Learning-based  (DL) models has proven their high efficiency for a wide range of Western languages.  In contrast, Arabic DL-based multimodal sentiment analysis (MSA)  is still in its infantile stage due, mainly, to the lack of standard datasets.
In this paper, our investigation is twofold. First, we design a pipeline that helps building our Arabic Multimodal dataset leveraging both state-of-the-art transformers and  feature extraction tools within word alignment techniques. Thereafter, we validate our dataset using state-of-the-art transformer-based model dealing with multimodality.
Despite the small size of the outcome dataset, experiments show that Arabic multimodality is very promising.

\end{abstract}


\section{Introduction}
\label{sec:intro}

The field of Multimodal Machine Learning (MML) has been growing rapidly over the past few decades, driven by the increasing availability of multimodal data and the need for more sophisticated and effective machine learning models. MML entails integrating and modeling data from various modalities (text, audio, image, video).


 Multimodal Sentiment Analysis (MSA), for instance, is an important and growing area of MML that aims to automatically determine the sentiment expressed in various modalities~\cite{soleymani2017survey}.  Early works in MSA deal with feature extraction and fusion processes in straightforward ways using standard machine learning algorithms~\cite{10.1145/2818346.2823317, PORIA201650}.  Over time, more complex methods, mainly deep learning models, were developed, such as CNN~\cite{Diwan2022}, RNN and its architectural variants~\cite{Tembhurne2021}, and Multimodal Multi-Utterance models~\cite{ABDU2021204}. 



Arabic MSA  is a promising area for academic research and practical applications due to the widespread use of the Arabic language and the increasing popularity of multimedia content. Furthermore, Arabic MSA is challenging due,  on the one hand,  to the complexity and the richness of the Arabic language, and on the other hand, to the significant cultural and linguistic variety of the Arab world. Therefore, it is still in its infancy~\cite{ghallab2020arabic}.

Despite these challenges, there has been some limited work in Arabic MSA that has shown promising results~\cite{najadat2018multimodal, alqarafi2019toward}. Hence, there is still much room for improvement in accuracy, efficiency, flexibility, and ability to handle diverse modalities.

This paper has two main investigations: 
\begin{enumerate}
    \item First, we design a pipeline  that facilitates the construction of a novel Arabic multimodal dataset. We accomplish this by leveraging state-of-the-art transformers and feature extraction tools alongside word alignment methods. 
    
    \item 
    To assess the effectiveness of our Arabic multimodal dataset, we employ cutting-edge transformer models that are intended to handle multimodality.  
\end{enumerate}

The remainder of this paper is structured as follows. 
Section~\ref{sec:BG} introduces some basic concepts concerning SA and MML required  to understand the rest of the paper.
Section~\ref{sec:RW} provides an overview of previous research on English and Arabic MSA.  
In Section~\ref{sec:Methode}, we describe the proposed methodology for multimodal dataset collection. We also present the models that were used to evaluate the designed dataset. Section~\ref{sec:Expeiments} deals with experiments and interpretation of the empirical findings.  
Finally, Section~\ref{sec:Conc} outlines the conclusions and future works.


\section{Preliminaries}
\label{sec:BG}
Before diving into the details of our approach, we start with the terminologies and background concepts that concern  Sentiment Analysis (SA)  and Multimodal Machine Learning (MML) elements: data representation, modality fusion methods, alignment, and pre-trained models.

\subsection{Sentiment Analysis}

Sentiment Analysis (SA), also referred to as opinion analysis, is the process of obtaining and examining the views, ideas, and perceptions of the public on a wide range of topics, products, subjects, and services. Corporations, governments, and people may all benefit from public opinion when gathering data and making choices based on it.~\cite{vinodhini2012sentiment}.  

Let us mention that the words \textit{emotion} and \textit{sentiment} are usually used interchangeably in daily life. While they are two different concepts. \textit{Emotion} is defined as a complex psychological state. There are six basic emotions, i.e., happiness, sadness, anger, fear, surprise, and disgust. This list is enriched by 
adding emotions such as pride, excitement, embarrassment, contempt, and shame. On the other hand, \textit{Sentiment} describes a mental attitude that is founded on emotion~\cite{yue2019survey}. Positive, neutral, and negative are the three fundamental polarities. The SA also makes reference to a polarity categorization. 
There are several methods for performing SA, including rule-based methods, machine learning-based methods, and hybrid approaches. Some popular machine learning-based methods include Naive Bayes, Support Vector Machines (SVM), and Deep Learning-based (DL)  models. However,  DL-based models have proved their efficiency as SOTA approaches.

Human natural perception refers to our ability to perceive and understand information from multiple modalities in a seamless and integrated way, such as seeing a picture and hearing a sound simultaneously to understand a concept. Multimodal SA aims to replicate this natural perception by combining information from multiple modalities (text, audio and image/video, and more) to improve the accuracy and efficiency of learning systems. 
  
\subsection{Multimodal Machine Learning}

Multimodal Machine Learning (MML) involves integrating and modeling multiple communicative modalities, Such as linguistic (text), acoustic (sound), and visual messages(image and video) of data, from a variety of diverse and interconnected sources~\cite{morency-etal-2022-tutorial}.
 By leveraging the strengths of different modalities, multimodal learning can help overcome the limitations of individual modalities and enhance overall learning performance.

Liang et al. proposed a taxonomy of six core features in MML: Modality representation, alignment, reasoning, generation, transference, and quantification~\cite{Liang2022} that are understudied in conventional unimodal machine learning. Considering their importance for our study, we focus on two features: i) Representation: where we focus mainly on which adequate representation is suitable for each modality and then how and when to fuse and integrate information from two or more modalities, effectively reducing the number of separate representations.
ii) Alignment:  Alignment between modalities is also challenging and involves identifying connections between modality elements.

\subsubsection{Fusion Methods}

 Basically, We have two main methods to make a fusion of modalities. The first is \textit{Early Fusion}, which happens when we mix the modalities before making decisions with concatenation, summation, or cross-attention mechanism. While the second is \textit{Late Fusion} method which makes a prediction based on each modality alone and then combines decisions to get a final prediction~\cite{xu2022multimodal}.
In our approach, We deploy an early fuse approach.

\subsubsection{Pre-trained Models}

The deployment of semantic and Deep Learning based approaches leads us generally to use some pre-trained models such as GloVe multimodal bi-transformer model (MMBT) models~\cite{pmlr-v139-radford21a}, CLIP~\cite{devlin2018bert},  BERT~\cite{pennington2014glove} and AraBERT~\cite{antoun2020arabert}. For the purpose of this paper, we have deployed AraBERT.   

BERT, which represents the basis of ArabBert, is a  Bidirectional Encoder Representation from Transformers developed by Google~\cite{devlin2018bert}. BERT large encompasses $24$ encoders with $16$ bidirectional self-attention heads trained from unlabeled data extracted from the BooksCorpus and English Wikipedia.

AraBERT~\cite{antoun2020arabert} is a pre-trained BERT transformer built for Arabic NLP tasks. It is trained on  $70$ million sentences, corresponding to $24$GB of Arabic text. AraBERT uses the same configuration as Bert. It has $12$ encoder blocks, $768$ hidden dimensions, $12$ attention heads, $512$ maximum sequence length, and  $110$M parameters.
\section{Related Work}
\label{sec:RW}

This section reviews the relevant studies in the field of multimodal sentiment analysis (MSA), including traditional, machine learning, and deep learning approaches for both English and Arabic languages.

\subsection{English Multimodal Sentiment Analysis}

The study of Zadeh~\cite{10.1145/2818346.2823317} is considered one of the pioneering works in the field of MSA. It is the first work to tackle the challenge of tri-modal (visual, audio, and textual features) sentiment analysis. The author creates a dataset of 47 videos from YouTube. Each input in the dataset was annotated with either a positive, negative, or neutral label. Moreover, the paper identifies specific subset of audio-visual features relevant to sentiment analysis and presents some instructions for integrating these features. In experiment, author uses the Hidden Markov Model (HMM) classifier. The findings demonstrate the promise of MSA despite the small size of the dataset and the straightforward text analysis method.

Poria et al.~\cite{PORIA201650} propose a novel methodology for performing MSA based on sentiment extraction from online videos. They deploy the dataset initially created by~\cite{10.1145/2818346.2823317}. The authors discuss features extracting process from various modalities (text, audio, and visual). These features are fused by incorporating different techniques (feature-level and decision-level). The authors used multiple supervised machine learning classifiers (Support Vector Machine (SVM), Extreme Learning Machine (ELM), Naive Bayes (NB), and Neural Networks) to validate their approach. Finally, a comparative study was carried out on the selected dataset, revealing that their proposed MSA system outperforms the current state-of-the-art systems. The best performance was achieved with Extreme Learning Machine (ELM) method.

The study in~\cite{Diwan2022} provides a detailed review that explores the applicability, challenges, and issues for textual, visual, and MSA using CNNs. Several enhancements have been proposed, such as combining  CNN and long short-term memory (LSTM) techniques.


Tembhurne and Diwan~\cite{Tembhurne2021} study the role of sequential deep neural networks in MSA. They thoroughly examined applicability, problems, issues, and methodologies for textual, visual, and MSA based on RNN and its architectural variants.

Recently, Abdu et al.~\cite{ABDU2021204} draw up a survey on MSA using deep learning. They have categorized $35$ cutting-edge models, recently suggested for the video sentiment analysis field, into eight  categories, based on the specific architecture employed in each model. After a detailed examination of the results, authors conclude that the \emph{Multimodal Multi-Utterance} based architecture is the most powerful in the task of MSA. 

Before concluding this section, we point out that the two transformer-based models known as the Multimodal transformer (Mult)~\cite{tsai2019multimodal} and LS-LSTM~\cite{6795963}  that we have used to evaluate our dataset are described in Section~\ref{sssec:Multimodal}.

\subsection{Arabic Multimodal Sentiment Analysis}
In contrast to the MSA studies made for the English language, the one for the Arabic language encompasses a limited number of works.

Najadat and Abushaqra~\cite{najadat2018multimodal} aim to address the issue of MSA for Arabic. They start by building their dataset from YouTube. They extract different features (linguistic, audio, and visual) from the collected videos. Also, they augment data using \emph{Weka} re-sampling option. For training and testing purposes, the authors annotate their dataset by positive, negative, and neutral polarities. In the experiment stage, the authors use different machine learning classifiers (Decision Trees, Support Vector Machine (SVM), k-Nearest Neighbor (KNN), Naive Bayes (NB), and Neural Networks). Obtained results reveal that the Neural Network classifier performs best when using only the audio modality. However, obtained results can be enhanced by feeding the dataset with more features.

 In their paper~\cite{alqarafi2019toward}, Alqarafi et al. try to tackle the problem of sentiment analysis in online opinion videos for modern standard Arabic. They begin by constructing their Arabic Multimodal Dataset (AMMD) from $40$ different YouTube videos. First, they used the extracted features (text, video) to feed their dataset. After that, they add metadata about the videos, including audio, transcription, visual motions, and sentiment polarities. Authors use, to conduct experiments, the Support Vector Machine (SVM) classifier. Despite the limited size of the dataset, the experimental results demonstrate the validity of the constructed dataset. Additionally, the results indicate that for several sentiment analysis tasks, including subjectivity and polarity classifications, the fusion of different features (utterance, visual) improves the performance compared to using utterance features alone.

\section{Methodology}
\label{sec:Methode}

In order to tackle the problem of Arabic Multimodal Sentiment Analysis \emph{AMSA}, in our study,  we conduct two main investigations. 

First,  we design a pipeline that eases building  Multimodal dataset for sentiment analysis that respects dataset collection engineering and harnesses transformers and SOTA feature extraction tools(~\ref{sec:PIPELINE}).  

Second, we assess our built dataset using SOTA  transformer-based models that deal with multimodality. The transformers are chosen to leverage inherent semantics while multimodality is deployed to improve the sentiment learner (Section ~\ref{sssec:Multimodal}).

\subsection{Multimodal Dataset collection Methodology}
\label{sec:PIPELINE}

As  mentioned above, our  targeted multimodal dataset for Arabic Sentiment Analysis, involving  dataset  collection engineering,    aims to leverage transformers and SOTA feature extraction techniques. Indeed, we have proposed this generic pipe to build the Multimodal dataset:

\begin{enumerate}
    \item Data Inventory, Collection, and Preprocessing.
    \item Annotation. 
    \item Data representation. 
\end{enumerate}

Our methodology is inspired by both  MOSEI~\cite{zadeh2018multimodal} and  CMU-MOSI~\cite{zadeh2016mosi} dataset-building processes while taking into account Arabic specificities.

\subsubsection{Data Inventory, Collection, and  Preprocessing} 

We rely mainly on videos on Youtube and Social Media platforms that include various information about the videos, such as audio, visual gestures, metadata, and probably transcripts. 
First, we have to identify sources that guarantee encompassing subjective information passages such as video-Bloggers, political analysts, and influencers channels. We also rely on some  Tv' Talk Shows. To get a large size of video we automatically draw some search lists and API to ease scraping information and its related metadata.    

For our targeted NLP task, pre-processing the collected
video include objective segment removal,  speech extraction, text extraction, and video/audio segmentation. All these processing are semi-automatic, using open source tools and quiet manual intervention. 

\subsubsection{Annotation}  
Annotating the polarity of video segments, as well as their associated text and speech, is the most challenging and resource-intensive task in our work. It requires a significant amount of time and resources to accurately annotate. We have opted to rely on manual crowdsourcing and manual annotation through a homemade platform. A guideline is devised in order to uniform the annotation.  In this step, we use the classic polarities $[-1,0,1]$ for negative, neutral, and positive sentiments, respectively. 
 %
 The annotation evaluation is performed through the  standard automatic Inter-Annotator Agreement method.

\subsubsection{Data Representation}
The three targeted modalities are represented in such a way that they exhibit more information on the inherent sentiment.  

\vspace{0.5em}
\noindent \textbf{Text}

Concerning the text, one can use either word-embedding or pretrained transformers. However, the latter allows learning contextual relationships between words in a sentence through a bidirectional attention mechanism. This means that we represent words taking into account both the left and right context of each word in a sentence, giving it a more comprehensive understanding of the semantic meaning and providing more accurate representations of textual modality. 
 
\vspace{0.5em}
\noindent \textbf{Visual Features}

The combination of body gestures and facial features can convey a more nuanced range of sentiments and emotions. 

Body gestures refer to physical characteristics of a person's body, such as Open/crossed arms, nodding, shaking head, and Shrugging shoulders.   
%
Facial features refer to the characteristics of a face, such as facial expressions and movements, that are used to represent emotions and sentiments of an individual. These features include commonly: smiling, frowning, raised eyebrows, squinted eyes, lip biting, tears, and blushing. 

For this version of our pipe,  we have opted to rely on facial features as they are the most commonly used features. In addition, they are also easier to capture from videos compared with body gestures.  
In fact, facial features can be extracted using computer vision techniques. It may include measurements of facial landmarks, facial action units, and head movements.
The  main descriptors we extract are: 

\begin{itemize}
 \item Action Units (AUs): these are facial muscle movements that are associated with various facial expressions such as brow raise, lip stretch, and eye closure.

\item Head pose: it estimates the orientation of the head in three dimensions, including pitch, yaw, and roll.

\item Eye gaze: it captures the direction of the eye gaze, including the location of the gaze and the direction of the gaze vector.

\item Facial Action Coding System (FACS): FACS is a system that describes facial expressions based on AUs.

\item Facial symmetry: it informs about the symmetry of the face by comparing the left and right sides of the face.
\end{itemize}

\vspace{0.5em}
\noindent \textbf{Acoustic Features}

The speech extracted from the video can be characterized at different levels acoustic, phonology, or prosody. The acoustic features are the most effective ones as they are language-independent since they rely on the physical features of the signal. However, also prosody features are essential as they capture features related to the emotion related to the speaker's speech.   These features are commonly used in speech recognition, speaker identification, and sentiment and emotion recognition systems. 
In our study, we  extract those main features:

\begin{itemize}
    \item Mel-Frequency cepstral coefficients (MFCCs): A set of coefficients that represent the spectral envelope of a speech \emph{signal}.
    \item Prosody: These include fundamental frequency (F0), speaking rate, and energy.
   \item Voice quality: These include jitter, shimmer, harmonic-to-noise ratio (HNR), and glottal waveform features.
   \item Emotion-related: These include pitch slope, pitch variance, and various modulation features.
   \item Spectral: These include spectral centroid, spectral flux, and spectral roll-off.
  \item Formant features: These include the first three formants, which are resonant frequencies of the vocal tract.
  \item Timing features: These include various measures of speech timing, such as pause duration and speech rate.

\end{itemize}

\subsubsection{Alignment Techniques}

One of the crucial aspects of the MML is the alignment of multimodal data, which involves synchronizing the different modalities, such as text, audio, and video. This is typically done by aligning the timestamps of each modality and mapping them onto a common timeline.

\noindent Considering text as an important modality, we use, in our work, two stages in achieving this alignment. In the first stage, we perform Text and Audio alignment, where data are aligned at the word level. In the second stage, we perform video and text alignment. Thus, we get a  global alignment over the text common modality. For both stages, we use forced alignment techniques. 
 
\vspace{0.5em}
\noindent \textbf{Forced Alignment Text-Audio}

Within this alignment, a transcript is synchronized with an audio recording by mapping each speech segment to its corresponding words. 
This process of forced alignment typically involves breaking down the audio and transcript into smaller segments and using algorithms to compare the speech and text segments to determine their correspondence. The algorithms consider various factors, such as speech timing, pronunciation of words, and speech sounds. After the forced alignment process, we get a time-stamped representation of speech. 

\vspace{0.5em}
\noindent \textbf{Pivot-Based Multimodal Alignment}

 For effective handling of multimodal time series data featuring multiple views at different frequencies,  it is crucial to align them to a designated "pivot" modality, which is typically done through textual modality. This involves grouping feature vectors from other modalities into bins based on the timestamps of the pivot modality and then applying a specific processing function, known as the "collapse function", to each bin. This function, often a pooling function, merges multiple feature vectors from another modality into a single vector, resulting in sequences of equal lengths across all modalities (matching the length of the pivot modality) in all-time series.

\subsection{Models}
\label{sssec:Multimodal}
  In this section, we provide a detailed explanation of the selected models used to validate our dataset. Our  explanation includes a discussion of crucial multimodal learning (MML) techniques used in these models, such as fusion, modeling, and alignments.
    
  \noindent The initial state-of-the-art model, known as the Multimodal transformer (Mult)\cite{tsai2019multimodal}, is a transformer-based model that deploys   an attention mechanism. It  allows each element of the input sequence $X_i$ to attend to all the other elements, resulting in a new weighted sequence $\widehat{X}_i$. This process is referred to as self-attention, as it enables the elements to focus on the most relevant ones, $i$ representing the modality among \{text, video, audio\}.
  
  Mult integrates these $\widehat{X}_t$,~$\widehat{X}_a$, and~$\widehat{X}_v$ by utilizing a pairwise feed-forward approach.
This is achieved through the implementation of a deep Cross Attention Block (CAB), which is defined below, 
where each pair-modality is fed into two CABs that alternate between Query, Key, and Values matrices. 
Before making predictions, the Multimodal Transformer model concatenates the $Z_{i \to j}$ and $Z_{k \to j}$ matrices obtained from the CABs, where $i,~j,$~and~$k$  represent distinct modalities. This process results in the 
$Z_{i}$~matrix. As previously mentioned, the prediction classifier is applied to the concatenated or summed $\widehat{Y}$, which is composed of the  $Z_t$,$Z_a$, and $Z_v$ matrices 
where $t$,~$a$,~and $v$ are referred to text, audio, and video respectively.
  \subsubsection*{Cross Attention Block (CAB)}
In order to implement the fusion from multiple modalities, Cross-modal allows the model to focus on relevant information from each modality and weigh the contribution of different modalities in the final prediction. 
We can express CAB in the mathematical formulation below. Let us consider an MML model with two  modalities: $X, Y$. The query, key, and value matrices for each modality can be represented as follows:
\begin{equation}
\label{eqn:regular_att0}
 Q_X = W_X^Q X, \ \ \ K_X = W_X^KX,\ \ \  V_X = W_X^VX
\end{equation}
\begin{equation}
\label{eqn:regular_att1}
CAB_{x\to y}= Softmax(\frac{Q_x . K_y^{T}}{\sqrt{d_k}}) . V_y  =Z_{x\to y},
\end{equation}

where $W_X^Q$, $W_X^K$, $W_X^V$, $W_Y^Q$, $W_Y^K$, $W_Y^V$ are the weight matrices for the query, key, and value computations for each modality.
 and $Q,K,V$ $\in \mathbb{R}^{l\star d}$, $d$ is the modality dimension, $l$ is the length of input token $X$.

\begin{figure}
    \centering
    \includegraphics[scale=0.475]{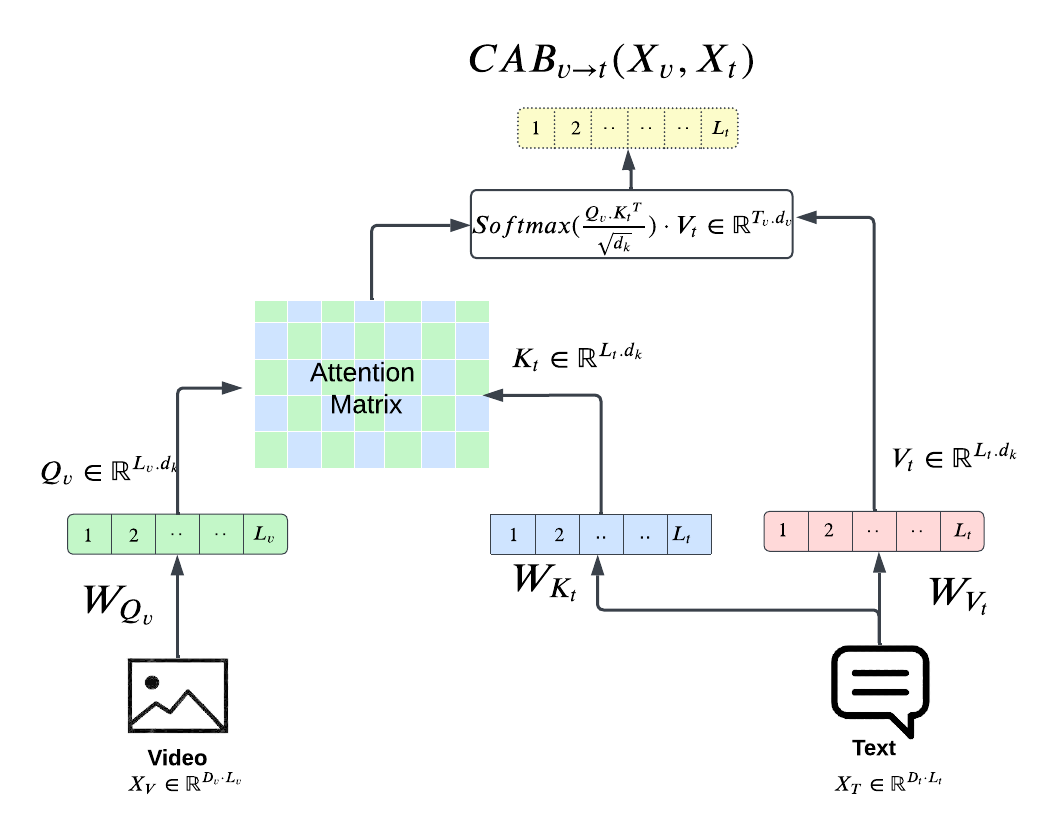}
    \caption{Cross-modal with CAB lead model to integrate input modalities. In this illustration, CAB combines video and text information ($X_v$ and $X_t$) through an attention mechanism.}
    \label{fig:cross}
\end{figure}

\noindent Fig.~\ref{fig:cross} illustrates the cross attention block.  The main purpose of using these two sub-layers before and after CAB layers is to make our model focus only on dependent features.

In light of Equations~(\ref{eqn:regular_att0}) and~(\ref{eqn:regular_att1}) described above, we can express for each modality latent representation $Z_i$ by Equation (\ref{eqn:regular_att3}).

\begin{equation}
\label{eqn:regular_att3}
Z_x =   \left [CAB_{y\to x} ; CAB_{k\to x} ]\right. 
\end{equation}

where $x, y,$ and $k$ are all possible modalities.\\
The output $\widehat{Y}$ of Mult model by Equation~(\ref{eqn:regular_att4}) as follows:
\begin{equation}
\label{eqn:regular_att4}
\widehat{Y} = \sum \left [Z_{t} ; Z_{a} ; Z_{v} ] \right. 
\end{equation}

In another hand, we select another deep learning model using another fusing approach. This model consists of three separate Long Short-Term Memory (LSTM) networks\cite{6795963}, one for each modality (text (t), visual (v), and acoustic (a)).  Here we use a late-fusion  where different modalities are processed separately to obtain their respective feature representations, and then these features are combined using a fusion mechanism to make the final prediction. In the first stage, the model takes as input three modalities $X_{\left\{ t,a,v \right\} }$  and extracts features $h1_i, h2_i$ from each of them using the corresponding two LSTMs with a normalization layer  between them as shown in equations bellow. 

    \begin{align*} o_t  &= \delta \left( W_o \times X_{\left\{ t/a/v \right\} }+h_{t-1}+b_o \right )\\
     i_t &= \sigma(W_i[ X_{\left\{ t/a/v \right\}}, h_{t-1}] + b_i) \\
      g_t &= \tanh(W_c[ X_{\left\{ t/a/v \right\}}, h_{t-1}] + b_c) 
    \end{align*}
    \begin{align*} O_{lstm1}(X)&= h1_t = o_t \odot \tanh(c_t)  \\
   O_{lstm2}(h1_t)&=h2_t =  o_t \odot \tanh(c_t)  
  \end{align*}

 These features are then concatenated and normalized, before being fed into a fully connected layer with a ReLU activation function and dropout regularization. Finally, the output is generated using another fully connected layer defined as: 
 
   $$Output\_model = O_{lstm2}\left(Norm\_Layer( O_{lstm1}(  X_{\left\{ t,a,v \right\}}))\right)$$ 
  $$Output\_model =Concat[h2_t;h1_t;h2_a;h1_a;h2_v;h1_v]$$

Where $x_t$ represents the input at time step $t$, $h_{t-1}$ represents the hidden state at the previous time step, $i_t$, $f_t$, and $o_t$ represent the input, forget, and output gates at time step $t$,  respectively. The symbol $\odot$ denotes element-wise multiplication, and $\sigma$ and $\tanh$ are the sigmoid and hyperbolic tangent activation functions, respectively.

\section{Experiments}
\label{sec:Expeiments}
In this section, first, we start by describing the details related to the implementation of our proposed Arabic Multimodal dataset pipe as well as the description of the collected dataset. 
In the second step, we empirically evaluate our built dataset through both Mult and LS-LSTM  models.

\subsection{Data collection }
\label{ssec:data-collection}

Following the above-designed pipe, our dataset is gathered from video-blogging' videos on mainly YouTube and some other social media platforms. The videos are retrieved and scraped automatically using a predefined list of keywords. In fact,  by means of this latter, we ensure the existence of subjective information related to Arabic content.

 \begin{figure*}[t]
    \centering
    \includegraphics[scale=0.6]{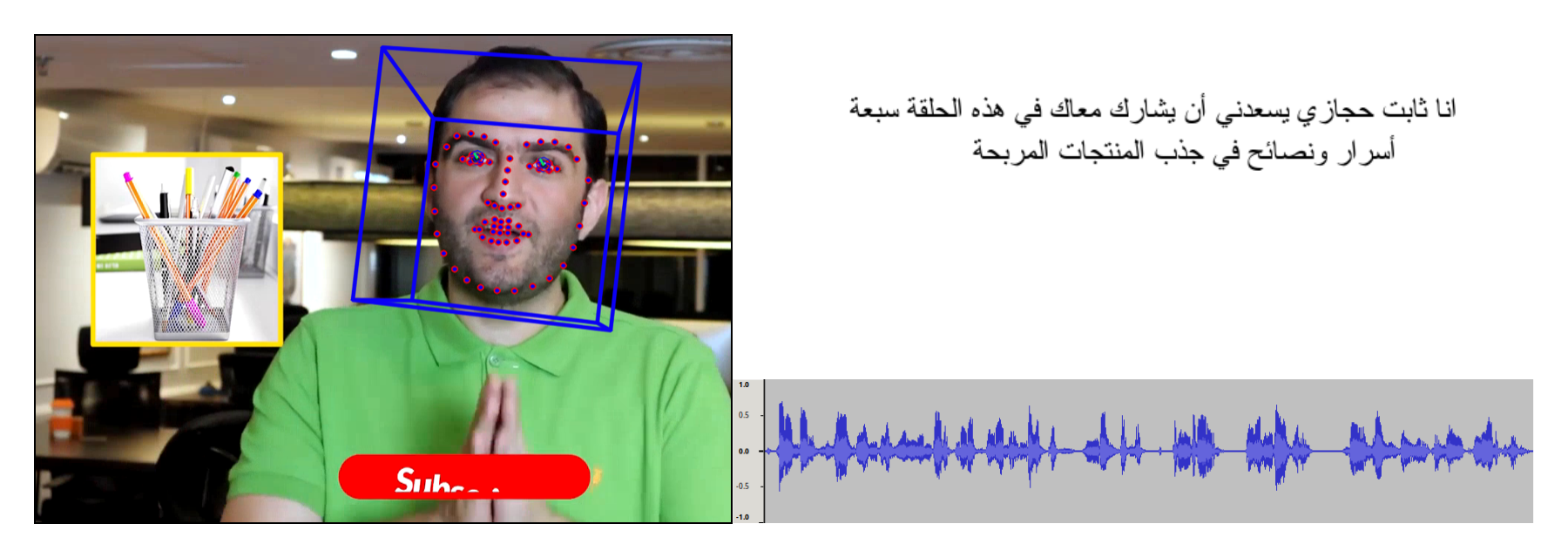}
    \caption{Illustration of an instance of our Dataset.}
    \label{fig:concat}
\end{figure*}

All videos have been checked manually to keep the most convenient ones for our study. Then using a homemade collaborative front-end tool~\footnote{\url{https://github.com/belgats/Arabic-Multimodal-Dataset/}}, We segmented each video by placing     \emph{start} and \emph{end} markers so that each video segment encompasses one subjective information.
Then we extracted from each video segment its related Arabic transcription and speech.  For those purposes, we use both  \empty{Klaam} tool~\footnote{\url{https://github.com/ARBML/klaam}} and \emph{Almufaragh} tool~\footnote{\url{https://almufaragh.com/}} for  Arabic speech recognition.  The automatically extracted transcripts are also checked manually to avoid and fix any transcription errors. The forced and pivot alignments are performed on the fly thanks to  Audacity tool~\footnote{\url{https://www.audacityteam.org/}}. 

Let us mention that word alignment is a challenging task. The quality of alignment can be negatively affected when dealing with speeches in which words are not fully enunciated by the speaker.

Concerning the annotation process, each segment is labeled by $5$ in lab annotators. A guideline is designed to reach more similar annotations. The Inter Agreement Annotator method is applied to assign a final label. 

The resulting three modalities (Video, text, and audio)  are then preprocessed to exhibit more information about their inherent sentiment, as described below.

Our method for extracting word vectors padded to max length from these transcripts is based on  AraBERT~\cite{antoun2020arabert} transformer. In fact, it is a BERT-based model that allows learning contextual relationships between words in a sentence through bidirectional attention mechanisms. That means that we represent words taking into account both the left and right context of each word in a sentence, giving it a more comprehensive understanding of the semantic meaning and providing more accurate representations of textual modality. Our text embeddings are in $768$ dimensional vector.

Concerning the visual features, we opted for the facial features. Thanks to OpenFace toolkit~\cite{7477553}, we extracted \nbVisualFeatures facial features belonging to those described above.  

The acoustic features are extracted using OpenSmile tool~\footnote{\url{https://OpenSmilehttps://www.audeering.com/research/opensmile}}.  We extracted \nbAcousticFeatures features described previously.

\begin{table}[tbp]
    \centering
    \caption{Dataset details.}
    \label{tab:datasetDetails}
    \begin{tabular}{L{4.2cm}R{2.6cm}}
    \toprule
   
    Number of videos & \nbVideo \\\midrule
    Total number of segments & 540\\\midrule
    Total number of subjective segments & \nbSegments \\\midrule
    Number of unique words & \nbWords  \\\midrule
    Total videos time & $02h:47min:27s$    \\\midrule
    Average length  of segments &   $17.46$ seconds 
    \\\midrule
     Number of Positive segment & 130 \\\midrule
     Number of Negative segment & 129 \\\midrule
     Number of Neutral segment & 59 \\\midrule
    Text embedding dimensions &  \nbArabBertFeautures \\\midrule
    Visual  Feature dimensions & \nbVisualFeatures \\\midrule
    Acoustic Feature & \nbAcousticFeatures 
    \\\midrule 
    Number of speakers & $23$
    \\\bottomrule 
    \end{tabular}   
\end{table}

Table~\ref{tab:datasetDetails} reports more details on the built  dataset.


After that, we construct data formatted as a dictionary of multiple computational sequences using CMU-multimodal SDK~\cite{zadeh2016mosi}.

A sample of our formatted dataset is available in~\footnote{\url{https://github.com/belgats/Arabic-Multimodal-Dataset/}}


\begin{figure*}
\centering
\includegraphics[scale=0.45]{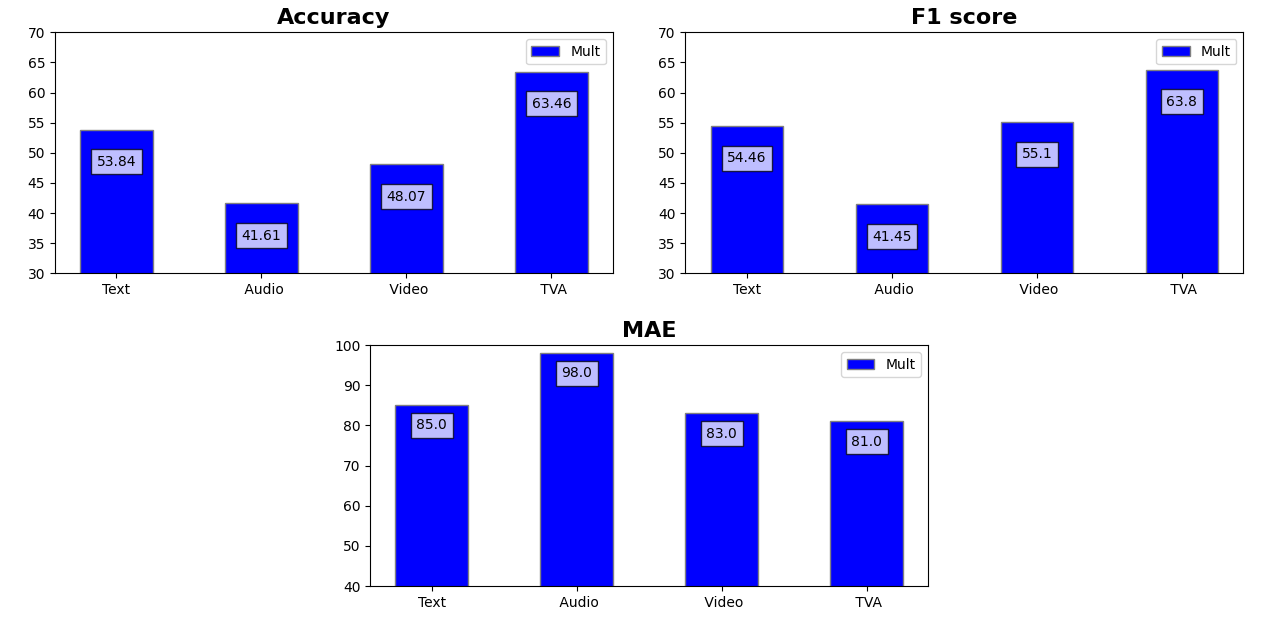}
\caption{Performances of Mult-based Models.}
    \label{fig:results-Mult}
\end{figure*}

\begin{figure*}
\centering

\includegraphics[scale=0.45]{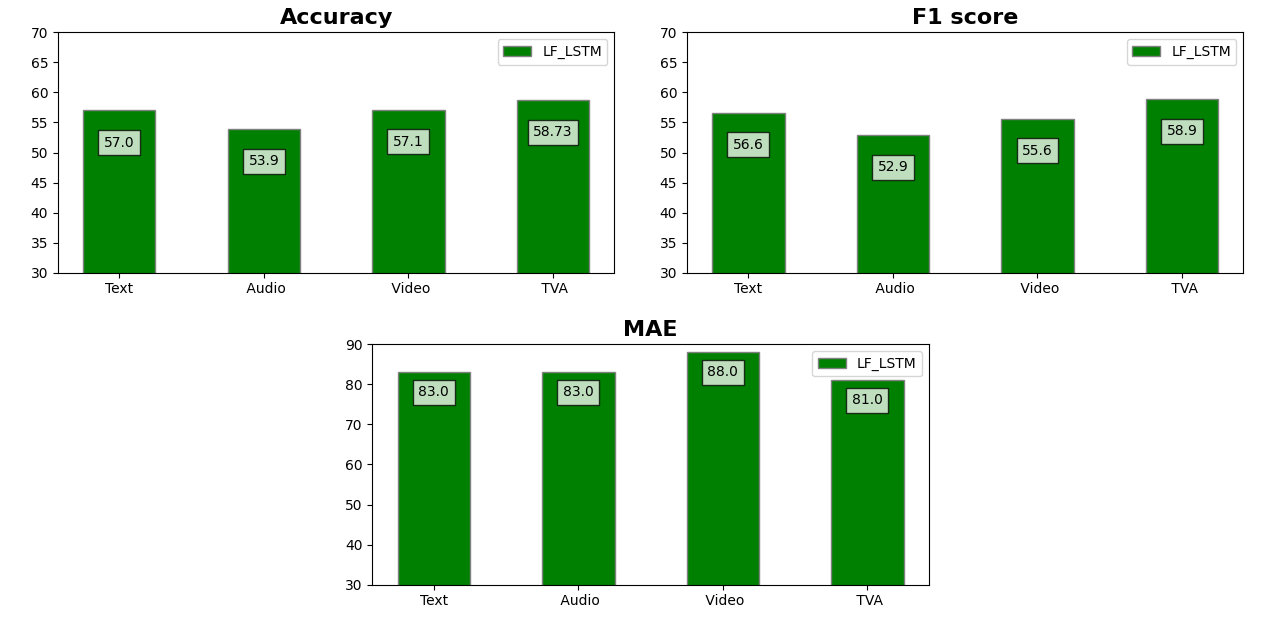}
\caption{Performances of LS-LSTM-based Models.}
    \label{fig:results-LSTM}
\end{figure*}

\subsection{Results and Discussion}
\label{ssec:Results}

The deployed models are measured through three metrics:  Accuracy, F1 score and  Mean Absolute Error (MAE). 
\noindent MAE tells us the mean absolute difference between predicted sentiment scores and the true sentiment scores. 
$$MAE = \frac{\sum_{i=1}^{n}\left| y_i - x_i\right|}{n}$$

\noindent Accuracy measures the proportion of true positives and true negatives out of total predictions.

\noindent F1 score measures the harmonic mean of precision and recall. 

 Figures~\ref{fig:results-Mult} and~\ref{fig:results-LSTM} report the performances of the Mult model (respectively LF-LSTM) using our Arabic Multimodal dataset in terms of Accuracy, F1, and MAE metrics. For each model, four variants are learned. Three uni-modal models considered  Text, Audio, and Video modalities alone. While TVA is the Multimodal that fuses the three modalities. 
 \\
 Let us mention that the uni-modal models for Mult are obtained by feeding the features of that specific modality with self-attention so that the CAB is replaced by a self-attention mechanism.


The results show that TVA  Mult-based learner outperforms the uni-modal models regarding all metrics. It improves the accuracy by $15.15$\%, $19.63$\%, and $18.22$\% for Text, Audio, and Video-based uni-models, respectively, while MAE is improved by $2$\% to  $4$\% compared to uni-modals.

Concerning the LS-LSTM-based models, the same result is observed. The  TVA  learner outperforms the uni-modal models regarding all metrics. However, with less improvement. Multimodality has enhanced the learner in terms of F1 score by $3.9$\%,  $10.19$\%, and $5.6$\% for Text, Audio, and Video-based uni-models, respectively. Furthermore, the  MAE has decreased by more than $8.64$\% for the text uni-modal model.

One can observe that the reached Multimodal based performances are not very high. F1 scores are about  $63,8$\% and $58,9$\% for Mult and LF-LSTM-based models, respectively. However, these models show their superiority compared to uni-modal models.
%

One could argue that these results are impacted by two factors. Firstly, the dataset size is relatively modest and needs to be expanded to ensure greater accuracy. Secondly, the alignment process is highly challenging. As previously mentioned, we encountered significant difficulties when dealing with speeches where the speaker swallowed words, which negatively affected the word alignment.

Another result to be underlined is the superiority of modalities early fusion (Mult) compared to the late fusion (LF-LSTM), one at least for our built dataset. This result is expected as it is also confirmed for other Languages' Multimodal models~\cite{tsai2019multimodal}. 



  \section{Conclusion and  Future Work}
 \label{sec:Conc}
In this paper, we have addressed the topic of multimodal sentiment analysis, which has the potential to revolutionize our understanding and analysis of human emotions, opening up new avenues for research and practical uses. However, to address the issue of the scarcity of Arabic multimodal datasets, we have developed a methodology for creating such a dataset. 
Subsequently, we assessed the effectiveness of our constructed dataset using state-of-the-art transformer models designed to handle multimodality. 

Despite the relatively small size of the constructed dataset, the findings show that considering multimodality is crucial for accurate Arabic sentiment analysis.

As further work, we intend to expand our Arabic multimodal dataset to meet the size requirements for deep learning algorithms. Furthermore, we conjecture that enhancing the alignment techniques used in the dataset can considerably improve the accuracy and effectiveness of sentiment analysis.





\bibliographystyle{IEEEtran}
\bibliography{My-references}

\end{document}